\documentclass{article}

\usepackage[final]{neurips_2021}

\usepackage[utf8]{inputenc} 
\usepackage[T1]{fontenc}    
\usepackage{hyperref}       
\usepackage{url}            
\usepackage{booktabs}       
\usepackage{amsfonts}       
\usepackage{nicefrac}       
\usepackage{microtype}      
\usepackage{xcolor}         

\usepackage{epsfig}
\usepackage{graphicx}
\usepackage{subfigure}
\usepackage{amsmath}
\usepackage{amssymb}
\usepackage{float}
\usepackage{stfloats}
\usepackage{underscore}
\usepackage{booktabs}
\usepackage{multirow}
\usepackage{caption}
\title{SizeNet: Object Recognition via Object Real Size-based Convolutional Networks}

\author{%
  Xiaofei Li, Zhong Dong \\
  Department of Automation\\
  Tsinghua University\\
  Beijing 100084, China \\
  \texttt{lixiaofei@tsinghua.edu.cn} \\
}

\begin{document}
\maketitle

\begin{abstract}
  Inspired by the conclusion that humans choose the visual cortex regions corresponding to the real size of an object to analyze its features when identifying objects in the real world, this paper presents a framework, SizeNet, which is based on both the real sizes and features of objects to solve object recognition problems. SizeNet was used for object recognition experiments on the homemade Rsize dataset, and was compared with the state-of-the-art methods AlexNet, VGG-16, Inception V3, Resnet-18, and DenseNet-121. The results showed that SizeNet provides much higher accuracy rates for object recognition than the other algorithms. SizeNet can solve the two problems of correctly recognizing objects with highly similar features but real sizes that are obviously different from each other, and correctly distinguishing a target object from interference objects whose real sizes are obviously different from the target object. This is because SizeNet recognizes objects based not only on their features, but also on their real size. The real size of an object can help exclude the interference object’s categories whose real size ranges do not match the real size of the object, which greatly reduces the object’s categories' number in the label set used for the downstream object recognition based on object features. SizeNet is of great significance for studying the interpretable computer vision. Our code and dataset will thus be made public.
\end{abstract}

\section{Introduction}
\label{in}

Recognizing objects in the real world in the same way as human beings is an important goal for artificial intelligence research [1]. At present, the computer vision object recognition research in the field of artificial intelligence is inspired by human vision system research results. Of these, various convolutional neural networks (CNN) [2-6] are the most advanced and widely used. CNN is a feature extraction and object recognition algorithm [7-10] which includes hierarchical structure, nonlinear activation, maximum pooling, and other operations. In addition, CNN is inspired by how the visual input of mammals can be visually recognized through hierarchical filtering and pooling operations of simple and complex cells in the V1 region [11]. Although CNN is inspired by the human visual neural network, the current CNN artificial neural network is a black box model and lacks interpretability.

How does the CNN neural network identify cats? One widely accepted answer is that it does so by examining the cat’s shape, and evidence for this hypothesis comes from visualization techniques such as DeconvNet [12]. However, some of the most important and widely used visualization techniques, such as DeconvNet, have recently been proven to be misleading [13]. Geirhos et al. found that for the deep neural network, a cat with an elephant’s skin is an elephant, but for humans, it is still a cat. Current deep learning techniques for object recognition, such as AlexNet, VGG-16, Googlenet, ResNet-50, ResNet-152, Densenet-121, and Squeezenet1_1, etc., mainly rely on the texture of objects rather than their shapes [14]. They also found that deep neural networks follow a shortcut learning strategy, that is, instead of correctly identifying and classifying foreground objects with labels, they learn to classify objects using a background, texture, or other such shortcuts (which are not obvious to humans) of foreground objects. This shortcut learning strategy makes the generalization ability of deep neural networks worse, lacks domain generality, and makes the recognition mechanism of neural networks incomprehensible [15]. Therefore, CNNs (model-based object recognition methods that only use feature information of an object’s image) face two problems. Firstly, they are unable to distinguish between objects whose features are highly similar, such as a real object, a model of an object, and a photograph of an object; secondly, they are unable to accurately recognize a target object when it is disturbed by a background or blocked by other objects. For example, a car with a clock texture is recognized as a clock, and a bear with a water bottle surface texture is recognized as a water bottle [14].

To solve the problems faced by CNN, some researchers have begun to recognize objects by considering the context of the objects in addition to the objects’ features. The difficulty of object recognition based on an object’s context is how to build a relationship between an object and its context. For example, desks and chairs often appear at the same time, but elephants and beds almost never appear at the same time; a plate is most often on a table, but is also likely to appear in other places. The relationships between objects and their contexts have strong and weak points which are obtained statistically from low-level features in an object’s image [16]. Although the context of an object is helpful for interpreting what it is, it greatly increases the time taken to recognize the object compared with CNN, and fails to solve the two major problems faced by CNN for two reasons. Firstly, contextual-based inference doesn’t work when different categories of objects whose features are highly similar but real sizes are very different appear in the same scenes normally, or appear in infrequent scenes in unusual situations; secondly, for situations whereby a target object has interference such as a background or other objects, contextual inference is even more powerless in most cases [17-20]. In addition, as an object can be photographed at different distances and angles, the object’s size in an image comes out of proportion, which invalidates the context-based computer vision model. Therefore, object recognition based on its context has many limitations, and contextual-based inference cannot solve the two major problems that CNN faces.

How can humans solve them solve those two problems with CNN? Visual commonsense tells us that human beings can use the real sizes of objects (the size of the overall shape or key parts of the object in the real physical world). Firstly, this allows objects with highly similar features to be easily recognized, such as cars, photos of cars, and models of cars; secondly, this allows target objects that are interfered with by the background and other objects to be correctly recognized. For example, accurately identify zebra crossings (target object) that are interfered with by cars, pedestrians, and other objects. At present, the research of human visual neural networks also confirms that the real size of an object is the main basis for choosing the visual cortex regions to analyze the features of an object for human visual object recognition [21-23]. The recognition of objects’ real sizes in human beings is an automatic [24-25] recognition process that first appears in infants, and is mediated by the dorsal visual pathway, ventral visual pathway, parietal cortex region, lateral occipital complex, and other brain regions [26-28]. This process occurs prior to object feature recognition [29-31]. It can be seen that the object recognition of human vision is based on not only an object’s features, but also on the real size of an object. Inspired by this, this article proposes SizeNet which is an object recognition method based on an object’s real size and features. At present, object recognition methods that use the real sizes of objects as an important basis for object recognition have not been reported in the field of computer vision.

\section{SizeNet}

\begin{figure}[t]
	\begin{center}
		\includegraphics[width=0.8\linewidth]{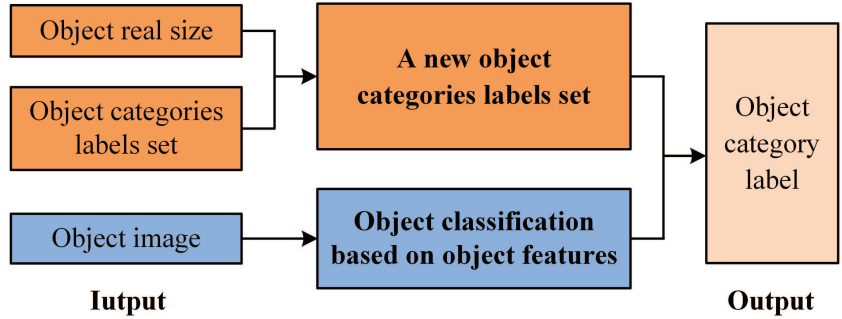}
		\setlength{\abovecaptionskip}{0pt}
		\setlength{\belowcaptionskip}{0pt}
	\end{center}
	\caption{Object recognition network of SizeNet.}
	\label{fig:fig1}
\end{figure}

Figure~\ref{fig:fig1} illustrates the end-to-end object recognition network used in SizeNet. The entire network can be split into two parts. The first part recognizes the real size of an object, and filters out the categories that match the real size of an object in the categories label set to generate a new categories label for the downstream object recognition, which based on an object’s features. The other part operates on an object’s features in an object image to propose and classify an object. The category label in the output layer of the object recognition algorithm based on an object’s features, which belongs to the new label set and has the highest matching probability, is output as the recognition result. The label set indicating the real size range of each category is manually calculated and annotated in advance. Due to the fact that the real size of each object in the same category is different, and the real size of the same object under different shooting angles is also different, the real size of each category label is therefore not a number, but a range.

\section{Experiments}

In order to verify whether SizeNet can effectively solve the two major problems faced by CNN, this paper created the dataset and designed the object recognition experiments.

\subsection{Dataset}
\label{dt}

\begin{figure*}[htp]
	\normalsize
	\centering
	\subfigure[The object images in the training sets of the Rsize dataset.]
	{	
		\label{fig:fig2a}
		\includegraphics[width=1.0\linewidth]{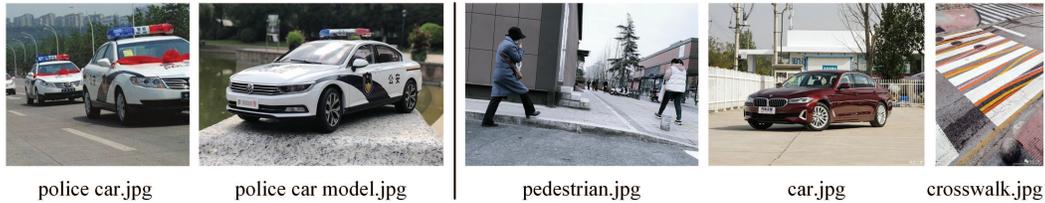}
	}
	\subfigure[The objects images marked with shooting distances in the test sets of the Rsize dataset.]
	{	
		\label{fig:fig2b}
		\includegraphics[width=1.0\linewidth]{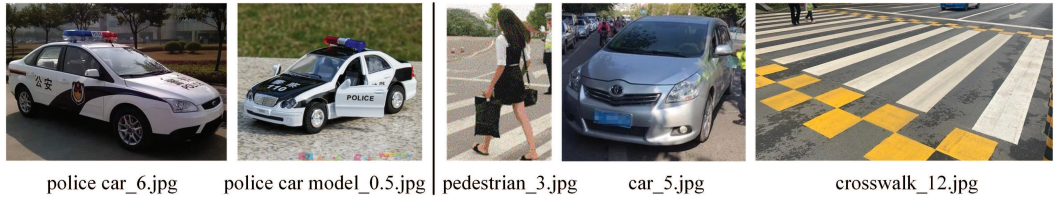}
	}
	\caption{The object images in the Rsize dataset. Figure~\ref{fig:fig2a} shows police car and a police car model in the training set of Rsize dataset 1, and some pedestrians, car, and a crosswalk in the training set of Rsize dataset 2. As the training images were used for training the object recognition algorithm based on object features, each image was not calibrated to the real sizes of the objects. Figure~\ref{fig:fig2b} shows a police car and a police car model marked with shooting distances in the test set of Rsize dataset 1, and a pedestrian, car, and a crosswalk marked with shooting distances in the test set of Rsize dataset 2.}
	\label{fig:fig2}
\end{figure*}

The self-made Rsize dataset for images, which annotated the real sizes of the objects, is shown in Table~\ref{table1} and Figure~\ref{fig:fig2}. Table~\ref{table1} shows the label set of the Rsize dataset. Each object category in the label set has a shooting distance range which was manually calculated in advance. For the two problems faced by CNN, Rsize dataset 1 in the Rsize dataset was used for objects with highly similar object features in recognition experiments, and Rsize dataset 2 was used for target objects that were interfered by backgrounds and other objects in recognition experiments.

Figure~\ref{fig:fig2} show some object images in the Rsize dataset. Figure~\ref{fig:fig2a} shows the images in the training set of the Rsize dataset. The image numbers of each object category in the training sets of the Rsize dataset are 350. Figure~\ref{fig:fig2b} shows the images in the test sets of the Rsize dataset. The image numbers of each object category in the test sets of the Rsize dataset are 100. Since SizeNet recognizes objects based on the real size of an object, each image in the test sets were calibrated to the real size of the object, and the real size of the object to be recognized in each image in the test sets was manually calculated in advance and marked in the name of each image. The number between the underscore "_" and ".jpg" in each image name denotes the real size of the object in question. These test sets use the shooting distance of the object (in meters) to indicate the real size of the object, because the real size of the object in the image was indicated by the shooting distance (the distance between the camera and the object to be recognized), and the size of the object in the image. In addition, the size of the object to be recognized in the image was made as large as the image size so that only the converted shooting distance could represent the real size of the object. 

\begin{table}[H]
	\setlength{\abovecaptionskip}{10pt}
	\setlength{\belowcaptionskip}{10pt}
	\normalsize
	\caption{The label set in the Rsize dataset, including Rsize dataset 1 and Rsize dataset 2. Rsize dataset 1 contains six categories of objects: Police cars (Pc), Police car models (Pcm), Fire trucks (Ft), Fire truck models (Ftm), Bullet trains (Bt), and Bullet train models (Btm); Rsize dataset 2 contains five categories of objects: Pedestrians (Pe), Cars (C), Crosswalks (Cw), Pillows (P), and Beds (B). Each object category in the label set has a shooting distance range (Sdr).}
	\label{table1}
	\centering
	\begin{tabular}{cccccccccccc}
		\toprule
		\multirow{2}{*}{} & \multicolumn{6}{c}{Rsize dataset 1} & \multicolumn{5}{c}{Rsize dataset 2} \\
		\cmidrule(r){2-7} \cmidrule(r){8-12}
		&  Pc   &  Pcm   &  Ft   &  Ftm   &  Bt   &  Btm
		&  Pe   &  C   &  Cw   &  P   &  B  \\
		\midrule
		Sdrs(m)    & 4-8 & 0.1-1 & 5-12 & 0.2-1 & 30-90 & 0.3-2 & 1-3.1 & 5-8 & 10-20 & 0.2-3 & 1.5-3.5  \\
		\bottomrule
	\end{tabular}
\end{table}

\subsection{Recognition architecture}
\label{ra}

\begin{figure*}[t]
	\begin{center}
		\includegraphics[width=0.9\linewidth]{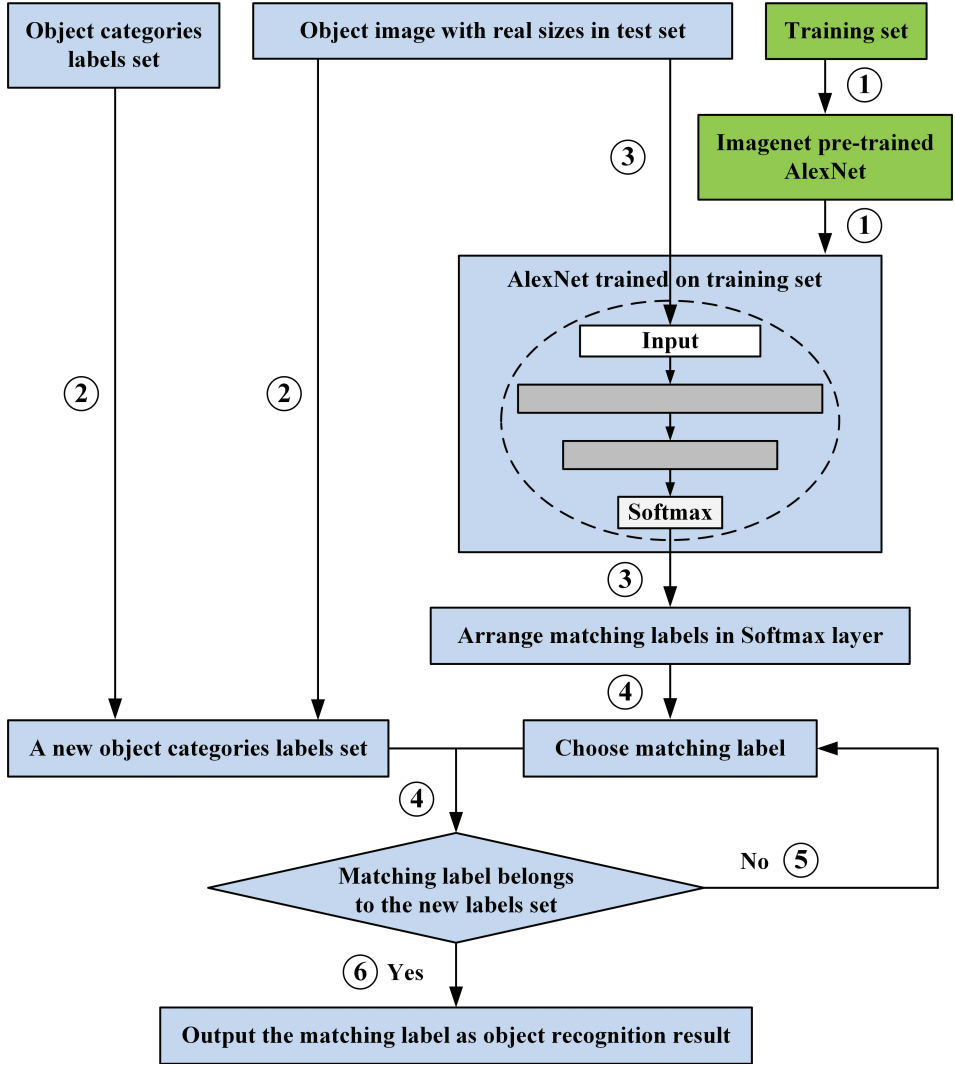}
		\setlength{\abovecaptionskip}{0pt}
		\setlength{\belowcaptionskip}{0pt}
	\end{center}
	\caption{The object recognition architecture of SizeNet on the Rsize dataset.}
	\label{fig:fig3}
\end{figure*}

Figure~\ref{fig:fig3} shows the object recognition architecture of SizeNet on the Rsize dataset. Step one, train the Imagenet pre-trained AlexNet which was used as the object recognition algorithm based on object features on training sets, and obtain the trained AlexNet. Step two, recognize the object image using the object’s real size in the test set one by one, and compare the object’s real size in the object image being recognized in the test set with the object real size ranges of all object categories in the object categories label set. This is done by choosing the object categories whose object real size ranges cover the real size of the object image being recognized to generate a new object categories label set. This process is downstream object recognition based on object features. Step three, recognize the object image being recognized using the trained AlexNet, and arrange the matching labels according to the matching probabilities from big to small in the Softmax layer when the input object image is passed to the Softmax layer. Step four, choose the matching labels in the Softmax layer from big to small, one by one, and determine whether the matching label belongs to the new label set or not. Step five, if the matching label does not belong to the new label set, i.e. “No”, then return to step four. Step six, if the matching label belongs to the new label set, i.e. “Yes”, then output the matching label as the object recognition result.

\subsection{Comparison with State-of-the-art Methods}
\label{cm}

In order to compare the object recognition performance of SizeNet, this paper used AlexNet, VGG-16, Inception V3, Resnet-18, and DenseNet-121 pre-trained on the ImageNet dataset (\url{https://download.pytorch.org/models}) to carry out the same object recognition experiments on the Rsize dataset. In order to study the influence of highly similar object features to carry out object recognition based on object features, this paper conducted object recognition experiments using SizeNet, AlexNet, VGG-16, Inception V3, Resnet-18, and DenseNet-121 on three datasets. One dataset includes objects such as police cars, fire trucks, and bullet trains; one includes object models such as police car models, fire truck models, and bullet train models; and the last one is the Rsize dataset 1, which includes both objects and object models such as police cars, fire trucks, bullet trains, police car models, fire truck models, and bullet train models. In order to study the influence of the background and other interfering objects to target object recognition, this paper conducted object recognition experiments on the Rsize dataset 2, whose training set includes the images which contain only the target objects, and a test set that includes the images which contain both the target objects and other interfering objects.

\section{Results}
\raggedbottom
Table~\ref{table2} shows the results of the object recognition experiments when using SizeNet, AlexNet, VGG-16, Inception V3, Resnet-18, and DenseNet-121 on the Rsize dataset 1. The results show that the average validation accuracies (Val acc. (\%)) and average test accuracies (Test acc. (\%)) for object recognition when using SizeNet, AlexNet, VGG-16, Inception V3, Resnet-18, and DenseNet-121 on the objects dataset or the objects models dataset in Rsize dataset 1 are all high, but on the Rsize dataset 1, which includes both objects and object models, the average test accuracies are all low (AlexNet 71.3\%, VGG-16 74.6\%, Inception V3 75.1\%, Resnet-18 74.8\%, and DenseNet-121 73.5\%). This was not the case for SizeNet however, which had a high test average accuracy of 98.0\%. The confusion matrix of real labels and prediction labels on the test set of Rsize dataset 1 using AlexNet is shown in Figure~\ref{fig:fig4}. From this, it can be concluded that AlexNet, VGG-16, Inception V3, Resnet-18, and DenseNet-121 cannot correctly recognize objects and object models whose features are highly similar, but SizeNet can correctly recognize the objects and object models based on their real sizes.

\begin{table}[t]
	\setlength{\abovecaptionskip}{10pt}
	\setlength{\belowcaptionskip}{10pt}
	\normalsize
	\caption{The results of object recognition experiments using SizeNet, AlexNet, VGG-16, Inception V3, Resnet-18 and DenseNet-121 on Rsize dataset 1.}
	\label{table2}
	\centering
	\begin{tabular}{cccccccccc}
		\toprule
		\multirow{2}{*}{} & \multicolumn{2}{c}{Objects} & \multicolumn{2}{c}{Object models} & \multicolumn{2}{c}{Rsize dataset 1} \\
		\cmidrule(r){2-3} \cmidrule(r){4-5} \cmidrule(r){6-7}
		&  Val acc.  &  Test acc.
		&  Val acc.  &  Test acc.
		&  Val acc.  &  Test acc.  \\
		\midrule
		SizeNet (ours)    & 98.0    & 100.0   & 95.1  & 98.0    & 86.5  & 98.0  \\
		AlexNet   & 97.5  & 99.0    & 93.1  & 95.6  & 84.6  & 71.3  \\
		VGG-16    & 96.8  & 99.1  & 94.2  & 98.6  & 85.8  & 74.6  \\
		Inception V3  & 93.1  & 99.0    & 89.9  & 96.3  & 80.1  & 75.1   \\
		Resnet-18   & 94.8  & 99.5  & 93.1  & 99.3  & 85.1  & 74.8 \\
		Densenet-121  & 96.2  & 99.6  & 92.8  & 98.1  & 85.0    & 73.5 \\
		\bottomrule
	\end{tabular}
\end{table}

\begin{figure}[htbp]
	\begin{center}
		\includegraphics[width=0.8\linewidth]{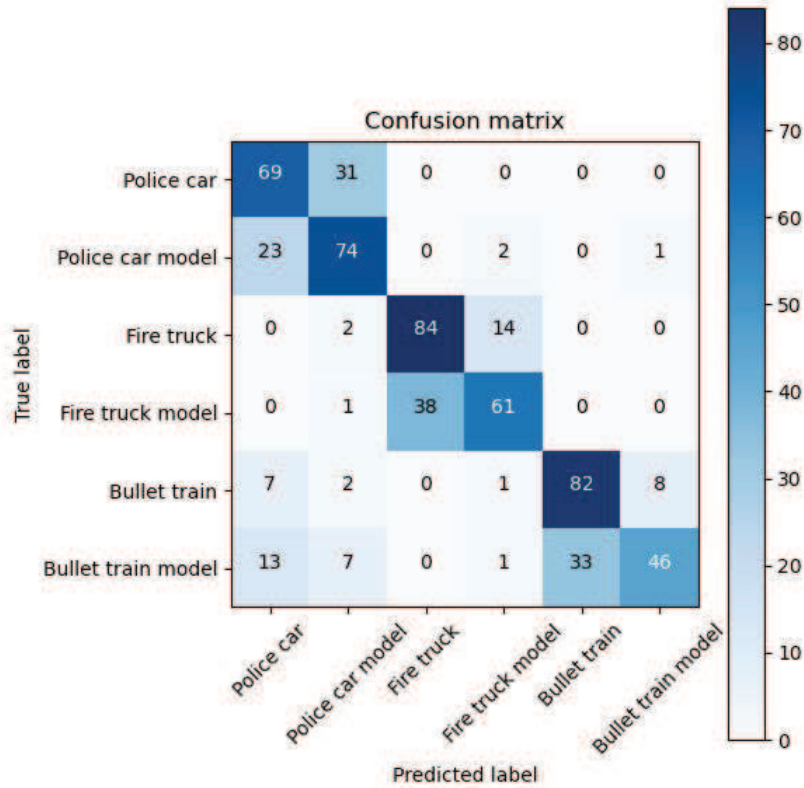}
		\setlength{\abovecaptionskip}{0pt}
		\setlength{\belowcaptionskip}{0pt}
	\end{center}
	\caption{The confusion matrix of real and prediction labels in the test set using AlexNet on Rsize dataset 1. It can be seen that 31 of the 100 images with police car real labels were misrecognized as police car models, and 23 of the 100 images with police car model real labels were misrecognized as police cars, and so on.}
	\label{fig:fig4}
\end{figure}

\begin{table}[t]
	\setlength{\abovecaptionskip}{10pt}
	\setlength{\belowcaptionskip}{10pt}
	\normalsize
	\caption{The results of object recognition experiments using SizeNet, AlexNet, VGG-16, Inception V3, Resnet-18 and DenseNet-121 on Rsize dataset 2.}
	\label{table3}
	\centering
	\begin{tabular}{ccc}
		\toprule
		& Val acc. & Test acc. \\
		\midrule
		Size-Net (ours) & 93.3  & 99.0 \\
		AlexNet & 93.8  & 71.2 \\
		VGG-16 & 92.0    & 69.4 \\
		Inception V3 & 90.1  & 65.2 \\
		Resnet-18 & 91.2  & 67.6 \\
		Densenet-121 & 92.5  & 67.8 \\
		\bottomrule
	\end{tabular}
\end{table}

\begin{figure}[t]
	\begin{center}
		\includegraphics[width=0.8\linewidth]{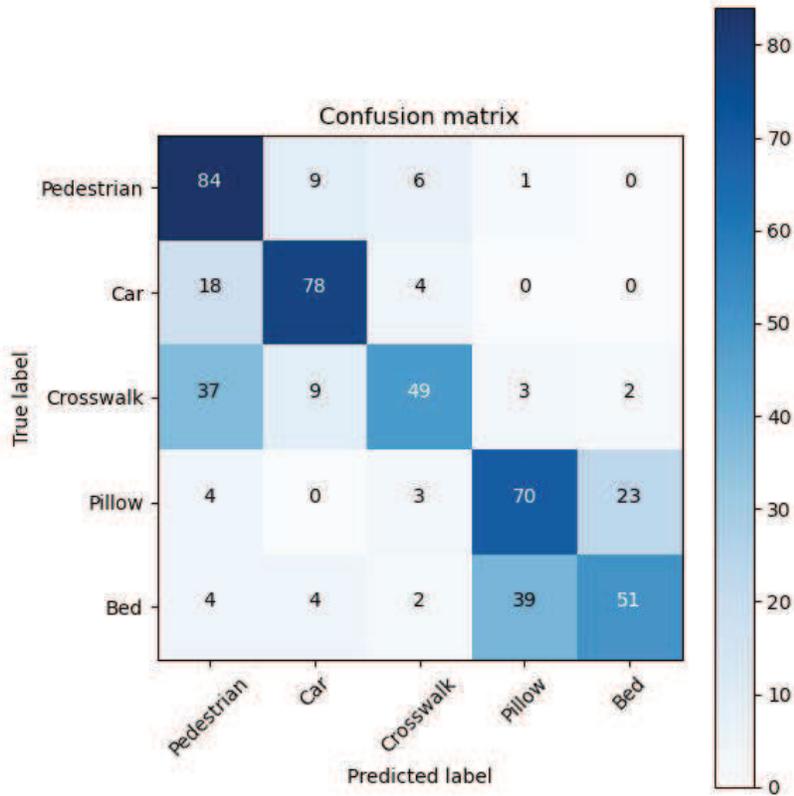}
		\setlength{\abovecaptionskip}{0pt}
		\setlength{\belowcaptionskip}{0pt}
	\end{center}
	\caption{The confusion matrix of real and predict labels in the test set using AlexNet on the Rsize dataset 2. It can be seen that 18 of the 100 images with car real labels were misrecognized as pedestrians, and 37 of the 100 images with crosswalk real labels were misrecognized as pedestrians; 23 of the 100 images with pillow real labels were misrecognized as beds, and 39 of the 100 images with bed real labels were misrecognized as pillows, and so on.}
	\label{fig:fig5}
\end{figure}

Table~\ref{table3} shows the results of object recognition experiments using SizeNet, AlexNet, VGG-16, Inception V3, Resnet-18, and DenseNet-121 on Rsize dataset 2. The results show that the average validation accuracies of object recognition using SizeNet, AlexNet, VGG-16, Inception V3, Resnet-18, and DenseNet-121 on the training set, which includes the images containing only target objects, are all high. However, on the test set which includes the images containing both target objects and other interfering objects, the average accuracies are all low (AlexNet 71.2\%, VGG-16 69.4\%, Inception V3 65.2\%, Resnet-18 67.6\% and DenseNet-121 67.8\%). But this was not the case for SizeNet, which obtained a high average test accuracy of 99.0\%. The confusion matrix of real and prediction labels on the test set of Rsize dataset 2 using AlexNet is shown in Figure~\ref{fig:fig5}. It can be concluded that AlexNet, VGG-16, Inception V3, Resnet-18, and DenseNet-121 cannot correctly recognize the target objects from other interfering objects whose real sizes are different, but SizeNet can correctly recognize and confirm the target objects from other interfering objects based on their real sizes.

This is because the SizeNet proposed in this paper recognizes the object is based on not only its features, but also on its real size. The real size of the object can avoid recognizing the object as a model or other objects, while their real sizes have a big difference.

\section{Discussion}
In order to solve the problems faced by CNN and context-based object recognition models, there are two issues that need to be addressed. Firstly, recognizing objects with highly similar object features; secondly, recognizing target objects that are interfered with by backgrounds and other objects. Inspired by the visual neural networks of humans, the SizeNet system proposed in this paper conducts object recognition based on the real sizes and features of objects. In addition, the Rsize dataset was used to address the two problems.

Object recognition experiments using SizeNet, AlexNet, VGG-16, Inception V3, Resnet-18, and DenseNet-121 were conducted on the Rsize dataset. The results show that the AlexNet, VGG-16, Inception V3, Resnet-18, and DenseNet-121 cannot correctly differentiate between police cars and the police car models, fire truck and fire truck models, and bullet train and bullet train models whose features are highly similar but real sizes are obviously different. This is because the AlexNet, VGG-16, Inception V3, Resnet-18, and DenseNet-121 only use the object features (shapes, textures, colors) to recognize the object, and do not consider the real sizes of the objects. The features of an object and an object model are highly similar, and therefore make it difficult to distinguish between them. The AlexNet, VGG-16, Inception V3, Resnet-18, and DenseNet-121 cannot correctly understand and distinguish target objects (such as pedestrians) from interference objects (such as crosswalks and cars), whose real sizes are different from the target objects. It is because the AlexNet, VGG-16, Inception V3, Resnet-18, and DenseNet-121 not only recognize the target object, but also the interference objects based on the object features, and cannot understand which object is the target object.

On the other hand, SizeNet can correctly recognize police cars and police car models, fire trucks and fire truck models, and bullet trains and bullet train models, and can also correctly understand and distinguish the target object from interference objects. This is because SizeNet not only uses the object’s features but also the real size of the object. Therefore, SizeNet does not correctly distinguish between the objects and object models from the highly similar features, but from the different real sizes, and correctly understands and distinguishes the target objects from the interference objects with the marked real sizes of the object images. SizeNet can solve the two problems faced by CNN and context-based object recognition models.

SizeNet can be integrated with any other object recognition algorithms based on object features, and can be used for image segmentation, object classification, and object detection. SizeNet has important theoretical and practical value for research into interpretable computer vision.

\section*{References}
\medskip
{
\small
[1] Yamins, Dlk\ \& Dicarlo, J. J.\ (2016) Using goal-driven deep learning models to understand sensory cortex. {\it Nature Neuroscience} {\bf 19}(3):356-365.

[2] Christian, S., Liu, W., Jia, Y., Pierre, S., Scott, R., Dragomir, A., Dumitru, E., Vincent, V.\ \& Andrew, R.\ (2015) Going deeper with convolutions. {\it 2015 IEEE Conference on Computer Vision and Pattern Recognition (CVPR)} {\bf 1}:1-9.

[3] Simonyan, K.\ \& Zisserman, A.\ (2014) Very Deep Convolutional Networks for Large-Scale Image Recognition. {\it Computer Science}.

[4] He, K.,  Zhang, X.,  Ren, S.\ \& Sun, J.\ (2016) Deep Residual Learning for Image Recognition. {\it 2016 IEEE Conference on Computer Vision and Pattern Recognition (CVPR)} {\bf 1}:770-778.

[5] Szegedy, C.,  Vanhoucke, V.,  Ioffe, S.,  Shlens, J.\ \& Wojna, Z.\ (2016) Rethinking the Inception Architecture for Computer Vision. {\it 2016 IEEE Conference on Computer Vision and Pattern Recognition (CVPR)} {\bf 1}:2818-2826.

[6] Huang, G.,  Liu, Z.,  Laurens, Vdm\ \& Weinberger, K. Q.\ (2017) Densely Connected Convolutional Networks. {\it 2017 IEEE Conference on Computer Vision and Pattern Recognition (CVPR)} {\bf 1}:2261-2269.

[7] Fukushima, K.\ (1980) Neocognitron: A self-organizing neural network model for a mechanism of pattern recognition unaffected by shift in position. {\it Biological Cybernetics} {\bf 36}(4):193-202.

[8] LeCun, Y., Boser, B., Denker, J. S., Henderson, D., Howard, R. E., Hubbard, W.\ \& Jackel, L. D.\ (1989) Backpropagation Applied to Handwritten Zip Code Recognition. {\it Neural Computation} {\bf 1}(4):541-551.

[9] Riesenhuber, M.\ \& Poggio, T.\ (1999) Hierarchical models of object recognition in cortex. {\it Nature Neuroscience} {\bf 2}(11):1019-1025.

[10] Krizhevsky, A., Sutskever, I.\ \& Hinton, G.\ (2012) ImageNet Classification with Deep Convolutional Neural Networks. {\it Advances in neural information processing systems} {\bf 25}(2):5249-5262.

[11] Hubel, D.H.\ \& Wiesel, T.N.\ (1959) Receptive fields of single neurones in the cat's striate cortex. {\it The Journal of Physiology} {\bf 148}(3):574-591.

[12] Zeiler, M.D.\ \& Fergus, R.\ (2014) Visualizing and Understanding Convolutional Neural Networks. {\it European Conference on Computer Vision} :818-833.

[13] Nie, W., Yang, Z.\ \& Patel, A.\ (2018) A Theoretical Explanation for Perplexing Behaviors of Backpropagation-based Visualizations. {\it Proceedings of the 35th International Conference on Machine Learning} {\bf 80}:3809-3818.

[14] Geirhos, R.,  Rubisch, P.,  Michaelis, C.,  Bethge, M.,  Wichmann, F. A.\ \& Brendel, W.\ (2019) ImageNet-trained CNNs are biased towards texture; increasing shape bias improves accuracy and robustness. {\it International Conference on Learning Representations}.

[15] Geirhos, R.,  Jacobsen, J. H.,  Michaelis, C.,  Zemel, R.,  Brendel, W.,  Bethge, M.\ \& Wichmann, F. A.\ (2020) Shortcut Learning in Deep Neural Networks. {\it Nature Machine Intelligence} {\bf 2}:665-673.

[16] Torralba, A.\ \& Sinha, P.\ (2001) Statistical context priming for object detection. {\it Proceedings of the IEEE International Conference on Computer Vision} {\bf 1}:763–770.

[17] Wolf, L.\ \& Bileschi, S.\ (2006) A Critical View of Context. {\it International Journal of Computer Vision} {\bf 69}(2):251-261.

[18] Rabinovich, A.,  Vedaldi, A.,  Galleguillos, C.,  Wiewiora, E.\ \& Belongie, S.\ (2007) Objects in context. {\it 2007 11th IEEE International Conference on Computer Vision} :1-8.

[19] Galleguillos, C.\ \& Belongie, S.\ (2010) Context based object categorization: A critical survey. {\it Computer Vision and Image Understanding} {\bf 114}(6):712-722.

[20] Zhao, H.,  Shi, J.,  Qi, X.,  Wang, X.\ \& Jia, J.\ (2017) Pyramid Scene Parsing Network. {\it 2017 IEEE Conference on Computer Vision and Pattern Recognition (CVPR)} {\bf 1}:6230-6239.

[21] Konkle, T.\ \& Oliva, A.\ (2012) A Real-World Size Organization of Object Responses in Occipitotemporal Cortex. {\it Neuron} {\bf 74}(6):1114-1124.

[22] Holler, D. E., Behrmann, M.\ \& Snow, J. C.\ (2019) Real-world size coding of solid objects, but not 2-D or 3-D images, in visual agnosia patients with bilateral ventral lesions. {\it Cortex} {\bf 119}:555-568.

[23] Holler, D. E., Fabbri, S.\ \& Snow, J. C.\ (2020) Object responses are highly malleable, rather than invariant, with changes in object appearance. {\it Scientific Reports} {\bf 10}(4654).

[24] Murray, S. O., Boyaci, H.\ \& Kersten, D.\ (2006) The representation of perceived angular size in human primary visual cortex. {\it Nature Neuroscience} {\bf 9}(3):429-434.

[25] Konkle, T.\ \& Oliva, A.\ (2012) A familiar-size Stroop effect: real-world size is an automatic property of object representation. {\it Journal of Experimental Psychology Human Perception \& Performance} {\bf 38}(3):561-569.

[26] Murata, A., Gallese, V., Luppino, G., Kaseda, M.\ \& Sakata, H.\ (2000) Selectivity for the Shape, Size, and Orientation of Objects for Grasping in Neurons of Monkey Parietal Area AIP. {\it Journal of Neurophysiology} {\bf 83}(5):2580–2601.

[27] Konen, C. S.\ \& Kastner, S.\ (2008) Two hierarchically organized neural systems for object information in human visual cortex. {\it Nature Neuroscience} {\bf 11}(2):224-231.

[28] Nishimura, M.,  Scherf, K,  Zachariou, V.,  Tarr, M\ \& Behrmann, M.\ (2014) Size Precedes View: Developmental Emergence of Invariant Object Representations in Lateral Occipital Complex. {\it Journal of Cognitive Neuroscience} {\bf 27}(3):1-18.

[29] Wilcox, T.\ (1999) Object individuation: infants' use of shape, size, pattern, and color. {\it Cognition} {\bf 72}(2):125-166.

[30] Long, B.,  Konkle, T.,  Cohen, M. A.\ \& Alvarez, G. A.\ (2016) Mid-level perceptual features distinguish objects of different real-world sizes. {\it Journal of Experimental Psychology General} {\bf 145}(1):95-109.

[31] Long, B\ \& Konkle, T.\ (2017) A familiar-size Stroop effect in the absence of basic-level recognition. {\it Cognition} {\bf 168}:234-242.
}

\end{document}